\pdfoutput=1
\documentclass[11pt]{article}

\usepackage{float}

\usepackage[]{EMNLP2023}

\usepackage{hyperref} % for clickable links
\usepackage{xurl}   

\usepackage{enumitem}

\usepackage{titlesec}

\usepackage{times}
\usepackage{latexsym}
\usepackage{tabularx}
\usepackage{tcolorbox}
\usepackage[T1]{fontenc}
\usepackage[utf8]{inputenc}

\usepackage{microtype}

\usepackage{inconsolata}

\usepackage{graphicx}
\title{NYUAD at ImageEval Shared Tasks: A Multimodal LLM for Audio-Visual Question Answering and Hallucination Detection in English and Modern Standard Arabic}

\author{Nouar AlDahoul \\
  Computer Science Department \\
   New York University \\
  Abu Dhabi, UAE \\
  \texttt{nouar.aldahoul@nyu.edu} \\\And
  Yasir Zaki \\
  Computer Science Department \\
   New York University \\
  Abu Dhabi, UAE \\
  \texttt{yasir.zaki@nyu.edu} }

\begin{document}
\maketitle
\begin{abstract}

\vspace{-5pt}

The increasing adoption of multimodal large language models (MLLMs) has enabled more natural interaction with visual and spoken information, creating new opportunities for intelligent audio-visual question answering (AVQA) and hallucination detection systems. AVQA extends traditional visual question answering by requiring models to jointly understand an input image and a spoken question to generate accurate, context-aware responses. Despite recent advances, AVQA and hallucination detection tasks remain challenging for modern standard Arabic (MSA) due to the limited availability of multimodal datasets and the need to reason across Arabic speech and text modalities alongside visual inputs.
In this paper, we present an MLLM-based framework for AVQA that takes an image and a spoken question as input and generates answers. In addition, we utilized this framework to detect hallucination in text-image pairs. The proposed approach is evaluated on the ImageEval 2026 Task (Ayn-VQA) benchmark, which focuses on culturally grounded Arabic AVQA. Experimental results demonstrate that the proposed framework effectively understands spoken queries, grounds their responses in the visual content, and produces accurate answers.
The results underscore the capacity and limitations of different LLM architectures for both English and MSA and under various conditions.
 
\end{abstract}

\section{Introduction}

\vspace{-5pt}

True human interaction is multi-sensory. As LLMs advance, bridging the gap between AI and human communication requires models that can simultaneously process text, audio, and visual information~\cite{aldahoul2025towards,imageeval-2026, alam2025everydaymmqa,imageeval-2026}.

Unlike traditional VQA, which relies on static text prompts, AVQA requires a model to dynamically align, synthesize, and reason across an input image and a spoken query to generate precise, context-aware textual responses.
By bridging the gap between diverse data streams, these models have moved beyond text-only interactions to enable more intuitive, human-like reasoning across visual and auditory inputs. This reasoning should also account for hallucinations that may distract the model from its main purpose.

While English-based multimodal systems have come a long way, adapting this technology to other languages is still a major hurdle~\cite{aldahoul2025nyuad1}. Arabic, in particular, faces unique challenges. First, there is a severe shortage of public datasets that connect Arabic speech with images. Second, the complex structure of MSA makes it very difficult for current models to connect Arabic audio with visual information~\cite{haouhat2025arabic}. To achieve equitable AI, we must use multimodal, dialect-sensitive datasets that reflect everyday cultural realities~\cite{aldahoul2025nyuad2,alam2025everydaymmqa}.

Building on existing works~\cite{mousi2026said}, this paper presents MLLM-based
framework for bilingual AVQA and hallucination detection, contributing to practical solutions that benefit the Arabic-speaking users. The first task~\cite{imageeval-2026} includes spoken questions about a given image, where the answers are visually present within the image itself. This requires models capable of processing and integrating speech and visual data simultaneously. To achieve this, we utilized several native multimodal models that directly process spoken questions to infer answers. Additionally, we evaluated a cascade pipeline approach that first transcribes the speech and then utilizes the extracted text alongside the image to determine the correct answers. The second task~\cite{imageeval-2026}includes an image with several statements, and the model must identify the incorrect ones. To address this identification, we utilized visual language models.

Our analyses and experiments center around the following research questions: 

\textbf{RQ1}: Can pre-trained multimodal Native LLMs effectively process spoken MSA and English queries and associated images to perform AVQA? 

\textbf{RQ2}: Can pre-trained cascade pipelines of multimodal LLMs effectively transcribe spoken MSA and English questions and generate correct answers from the pairs of transcripts and images? 

\textbf{RQ3}: Can pre-trained multimodal LLMs effectively detect hallucinations in VQA?

\textbf{RQ4}: Do multimodal LLMs that achieve high accuracy in English AVQA maintain the same level of performance in MSA equivalents?

\vspace{-5pt}

\section{Related Work}

\vspace{-5pt}

\subsection{AVQA Datasets}

\vspace{-3pt}

Due to the scarcity of large-scale Arabic datasets, prior studies have frequently translated English corpora into Arabic to generate synthetic data across various domains. Direct translations from English to Arabic often fail to capture the nuanced cultural and linguistic contexts, resulting in benchmarks that do not accurately assess a model's true capability in understanding natural Arabic text~\cite{eltahir2025autoarabic}

Several works have proposed large-scale English datasets of audio-visual question-answer pairs designed to advance multi-modal understanding~\cite{li2022learning,wu2025avqacl,radevski2026dave}. In the context of the Arabic language, researchers created the CAMEL-Bench benchmark dataset, which comprises around 29,036 questions and eight diverse domains. It targets the Arabic language, which represents a large population of over 400 million speakers~\cite{ghaboura2025camel}. 
In addition, recently, researchers have created OASIS, a large-scale, culturally grounded multimodal QA dataset featuring 920,000 images and 14.8 million text and speech pairs across English and Arabic dialects~\cite{alam2025everydaymmqa}.

\subsection{Native Multimodal Modeling}

Native multimodal modeling (NMM) is a single, unified AI brain that can see, hear, and read all at the same time~\cite{an2026toward}.
This approach allows the AI to process pixels, sound waves, and words in the exact same way, giving it a much deeper, more natural understanding of how the real world actually works~\cite{an2026toward}.

Recent breakthroughs in artificial intelligence have seen the rapid rise of models capable of cross-modal understanding. Several state-of-the-art models highlight this shift: First, Gemini 2.5 Flash\footnote{\url{https://docs.cloud.google.com/gemini-enterprise-agent-platform/models/gemini/2-5-flash-live-api}} is a natively multimodal AI model capable of processing and analyzing text, audio, image, and video inputs. Second, NVIDIA Nemotron 3 Nano Omni is an open omnimodal model designed to natively support audio inputs  alongside text, images, and video, powering enterprise-grade QA, summarization, transcription, and document intelligence workflows~\cite{nvidia2026nemotron3nanoomni}. Finally, Gemma 4\footnote{\url{https://ai.google.dev/gemma/docs/core/model_card_4}} includes natively multimodal models that can process raw audio waves and video frames directly inside their core neural network.

\vspace{-5pt}

\subsection{Non-Native Multimodal Modeling}

\vspace{-3pt}

In non-native AVQA, separate models are chained together, such as Whisper~\cite{radford2023robust} and Chirp-3\footnote{\url{https://docs.cloud.google.com/speech-to-text/docs/models/chirp-3}}, for speech to text (STT) and other language models for text processing alongside the image. As a result, the second model's reasoning base is entirely blind to the raw sensory signals.

On the other hand, researchers combined LLMs with encoders to achieve multimodal capabilities, such as video or visual and auditory content understanding. Video-LLaMA is a framework that enables LLMs to understand both the visual and auditory components of videos by connecting frozen, pre-trained encoders to a frozen LLM~\cite{zhang2023video}. In addition, Video-ChatGPT is a model designed for video-based conversation by merging a video-adapted visual encoder with an LLM~\cite{maaz2024video}.

\vspace{-7pt}
\section{Materials and Methods}
\vspace{-7pt}
This section provides an overview of the datasets used and outlines the methods employed. 

\subsection{Dataset Overview}

Here we describe the datasets proposed in the ArabicNLP ImageEval 2026 challenge~\cite{imageeval-2026} that we utilized to run our experiments. 

The first set of datasets is used in the first task that aims to build, evaluate, and compare multimodal systems for English and MSA AVQA. It includes 3,000 examples for training, 500 for validation, 500 for testing, and 1,000	for blind final ranking. The goal of using these datasets is to find an answer for spoken VQA, which includes an image and a spoken question with three spoken options. The solution should find the option index that is aligned with the image content.

The second set of datasets~\cite{mousi-etal-2026-correct} is utilized in the second task that aims to build, evaluate, and compare hallucination detectors in the task of English and MSA text-based statements aligned with images. It includes 3,000 examples for training, 500 for validation, 500 for testing, and 1,000 for blind final ranking. The goal of using these datasets is to find a true statement that is aligned with image content from a list of three statements. 

Figure~\ref{fig:examples1}, Figure~\ref{fig:examples2}, Figure~\ref{fig:examples3}, and Figure~\ref{fig:examples4} in Appendix A show examples from the AVQA and hallucination detection datasets in English and MSA.

\subsection{Methods}
\vspace{-3pt}

\begin{figure}[H]
    \centering
    \includegraphics[width=1\linewidth]{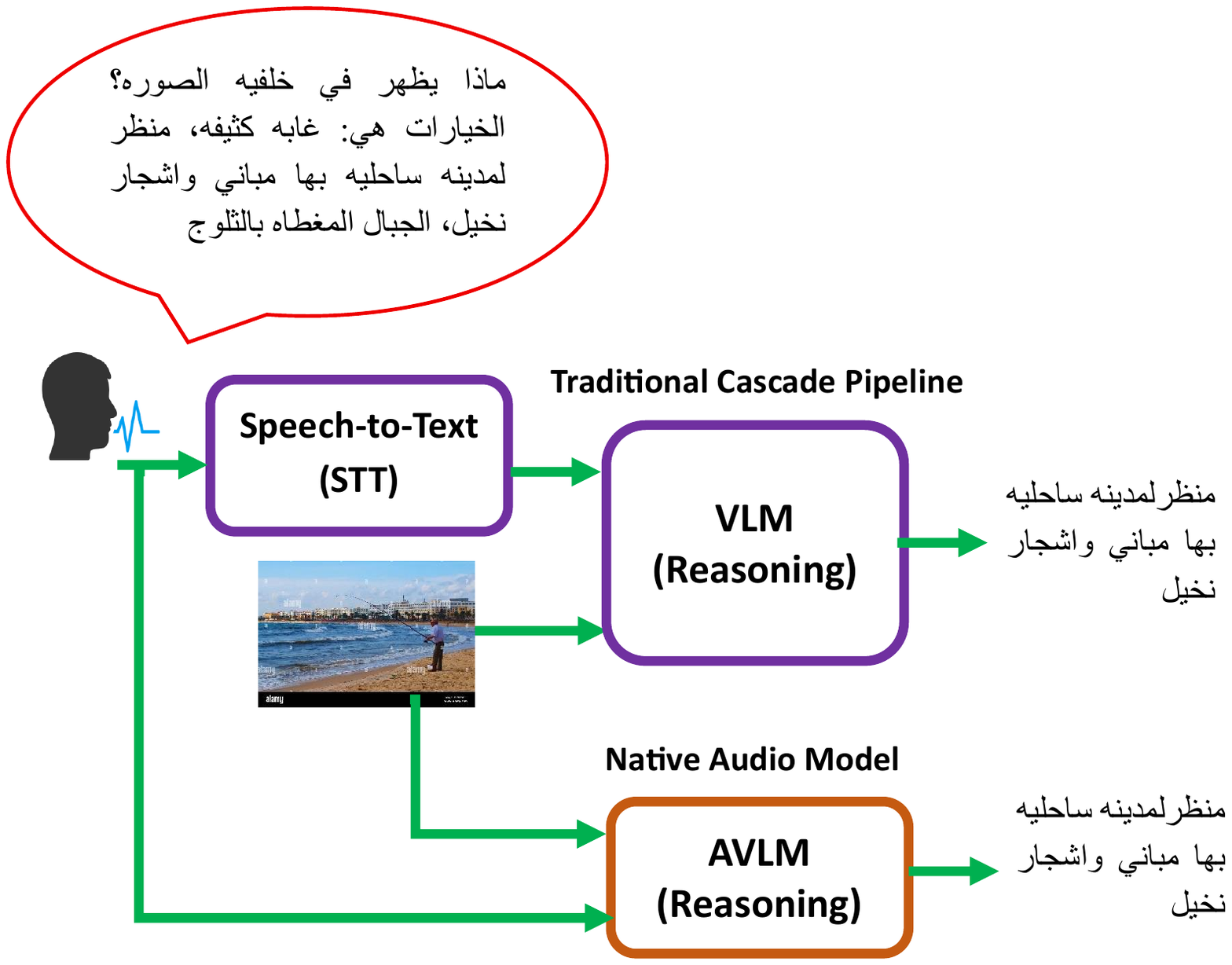}
    \caption{Overview of spoken visual question-answering approaches. The traditional cascade pipeline transcribes spoken MSA questions into text before feeding it into a VLM, whereas the native audio model directly integrates spoken question and image inputs into the VLM to generate the final response}
    \label{fig:diagram}
\end{figure}

For this task, we have evaluated several pre-trained native multimodal models to find the best solution that produced the highest accuracy and F1 score. We used OpenRouter APIs\footnote{\url{https://openrouter.ai/}} to run Gemini 3.5 Flash, Gemini 2.5 Flash, GPT-4o, Chirp-3, and Whisper-1 models. To run Gemma-4-12B, Gemma-4-4B, and Cohere-transcribe, we utilized GPU A100-40GB with the Hugging Face library. Figure~\ref{fig:diagram}  illustrates the traditional cascade pipeline (ASR + VLM) and the native audio models utilized in this work. The prompts used in this work are listed in Appendix A.

The first group of solutions includes audio-native language models such as Gemini 2.5 Flash, Gemma-4-4B, and Gemma-4-12B, which are audio LLMs that ingest speech directly via neural audio tokenizers or speech encoders rather than separate text automatic speech recognition (ASR) front-ends. In the first task, we utilized the previous three models with the first dataset to find the correct answer from a spoken question in English and MSA given alongside an image.  
Beyond basic ASR, Gemma-4-12B and Gemma-4-4B models can handle spoken question answering. While Gemma-4-4B has an audio encoder (alongside a vision encoder) to process raw audio before passing representations into the main language model, Gemma-4-12B is trained using a unified, encoder-free architecture that projects raw audio signals directly into the language model's transformer backbone without using a separate dedicated audio encoder\footnote{\url{https://developers.googleblog.com/gemma-4-12b-the-developer-guide/}}. On the other hand, Gemini 2.5 Flash, which is a native audio mode, utilizes a single model to handle audio directly and process audio at 32 tokens per second. It processes raw audio and images natively through unified multimodal architectures\footnote{\url{https://docs.cloud.google.com/gemini-enterprise-agent-platform/models/gemini/2-5-flash-live-api}}.

The second set of solutions is based on non-native multimodal models such as Chirp-3, Whisper-1, and Gemma-4-12B that work as ASR to generate transcripts in English or MSA. The transcripts are sent to other VLMs such as Gemini 3.5 Flash, GPT-4o, and Gemma-4-12B to find the correct answer. All three previous models support English and Arabic
\footnote{\url{https://docs.cloud.google.com/speech-to-text/docs/speech-to-text-supported-languages}}
\footnote{\url{https://github.com/openai/whisper\#available-models-and-languages}}
\footnote{\url{https://ai.google.dev/gemma/docs/core/model_card_4}}.

The third category of solutions focuses on multimodal reasoning using text-image pairs to identify the false statements among three provided options in English and MSA, evaluated alongside an image. Gemini 3.5 Flash and Gemma-4-12B models were employed for this task.

\section{Results and Discussion}
\vspace{-3pt}

\subsection{Development Phase}

\subsubsection*{Audio-Visual Question Answering}
Table~\ref{tab:task1} compares different native and non-native multimodal models for pairs of English-spoken questions alongside images. The results indicate that Gemini 2.5 Flash, as a native audio model, achieved the highest performance across all metrics (97.4\% accuracy and 97.4\% macro F1 score), outperforming other solutions, answering \textbf{RQ1}. In addition, Chirp-3 as ASR and Gemini 3.5 Flash as reasoning VLM achieved the second-highest performance, followed by Whisper-1 (ASR) + GPT-4o. However, smaller open-weight models like Gemma-4-4B lag in accuracy compared to frontier models and larger variants such as Gemma-4-12B, which strikes an optimal balance between a lower parameter count and strong performance (achieving 82.0\% as native audio and 85.6\% as ASR).

\vspace{-3pt}
\begin{table}[h]
\centering
\footnotesize
\begin{tabularx}{\columnwidth}{m{1.35cm}XXXX}
\hline
\textbf{Solution} & \textbf{Input} &  \textbf{Accuracy}  & \textbf{Macro F1-Score} \\ \hline
\textbf{Gemini 2.5 Flash} & Audio         & \textbf{0.974}           &\textbf{0.974}                          \\ 

\textbf{Native Gemma 4 12B}  & Audio   & 0.820            & 0.820             \\ 
\textbf{Native Gemma 4 4B}  & Audio   & 0.716            & 0.715            \\ 
\textbf{Cascade two Gemma 4 12B}   & Audio ->Transcript  & 0.856         & 0.869                         \\ 

\textbf{Chirp-3 + Gemini 3.5 Flash}      & Audio ->Transcript                   & 0.950         & 0.950                      \\ 
\textbf{Whisper-1 + GPT-4o} & Audio ->Transcript  & 0.906            & 0.908                                    \\ \hline
\end{tabularx}%
% }
\caption{\normalsize Performance metrics for  English audio-visual question-answering}

\label{tab:task1}
\end{table}

\vspace{-10pt}

Moving to MSA-spoken visual answering, Gemini 2.5 Flash exhibits extremely high performance when the input is an English-spoken question. However, its performance degraded largely by 19\% when the input is an MSA spoken question, answering \textbf{RQ4}. 

In addition, we found that cascaded ASR (Chirp-3 + Gemini 3.5 Flash) achieved the highest performance across all metrics (92.0\% accuracy, 92.11\% macro F1 score), outperforming end-to-end native multimodal solutions. 
This implies that converting MSA audio to text via a dedicated ASR engine (Chirp-3) before passing it to a VLM completely bypasses the semantic loss that native audio encoders currently suffer from.
Chirp-3 appears far better adapted to MSA phonetic structures, accents, and acoustic properties, leading to cleaner downstream transcriptions for the VLM, answering, \textbf{RQ2} and \textbf{RQ4}. 
While OpenAI Whisper-1 shows high performance with English-spoken questions, it struggles far more with MSA compared to Chirp-3.  

Feeding spoken MSA questions into Gemma 4 12B and 4B as single models results in low accuracy. Once the spoken MSA is discretized into standard Arabic text, Gemma 4's internal text-and-vision layers can process the visual task effectively without audio feature noise interfering with visual attention. This explains the jump in accuracy by 18\% in two Gemma-4-12B models (ASR + VLM). 

The model's bottleneck in native mode is not its inability to hear spoken MSA, but its inability to jointly align native audio representations with visual tokens during cross-modal reasoning.

\vspace{-3pt}
\begin{table}[h]
\centering
\footnotesize
\begin{tabularx}{\columnwidth}{m{1.35cm}XXXX}
\hline
\textbf{Solution} & \textbf{Input} &  \textbf{Accuracy}  & \textbf{Macro F1-Score} \\ \hline

\textbf{Gemini 2.5 Flash} & Audio         & 0.780           &0.772                          \\ 

\textbf{Native Gemma 4 12B}  & Audio   & 0.500           & 0.491           \\ 
\textbf{Native Gemma 4 4B}  & Audio   & 0.476            & 0.471           \\ 
\textbf{Cascade two Gemma 4 12B}   & Audio 
->Transcript  & 0.684         & 0.686          \\ 
\textbf{Chirp-3 + Gemini 3.5 Flash}      & Audio ->Transcript                   & \textbf{0.920}         & \textbf{0.921}                      \\ 
\textbf{Whisper-1 + GPT-4o} & Audio 
->Transcript & 0.794            & 0.791                                    \\ \hline
\end{tabularx}%
% }
\caption{\normalsize Performance metrics for  MSA audio-visual question-answering}
\label{tab:task2}
\end{table}

\subsubsection*{Error Analysis in MSA Transcripts}

\begin{figure}[H]
    \centering
   \includegraphics[width=1\linewidth]{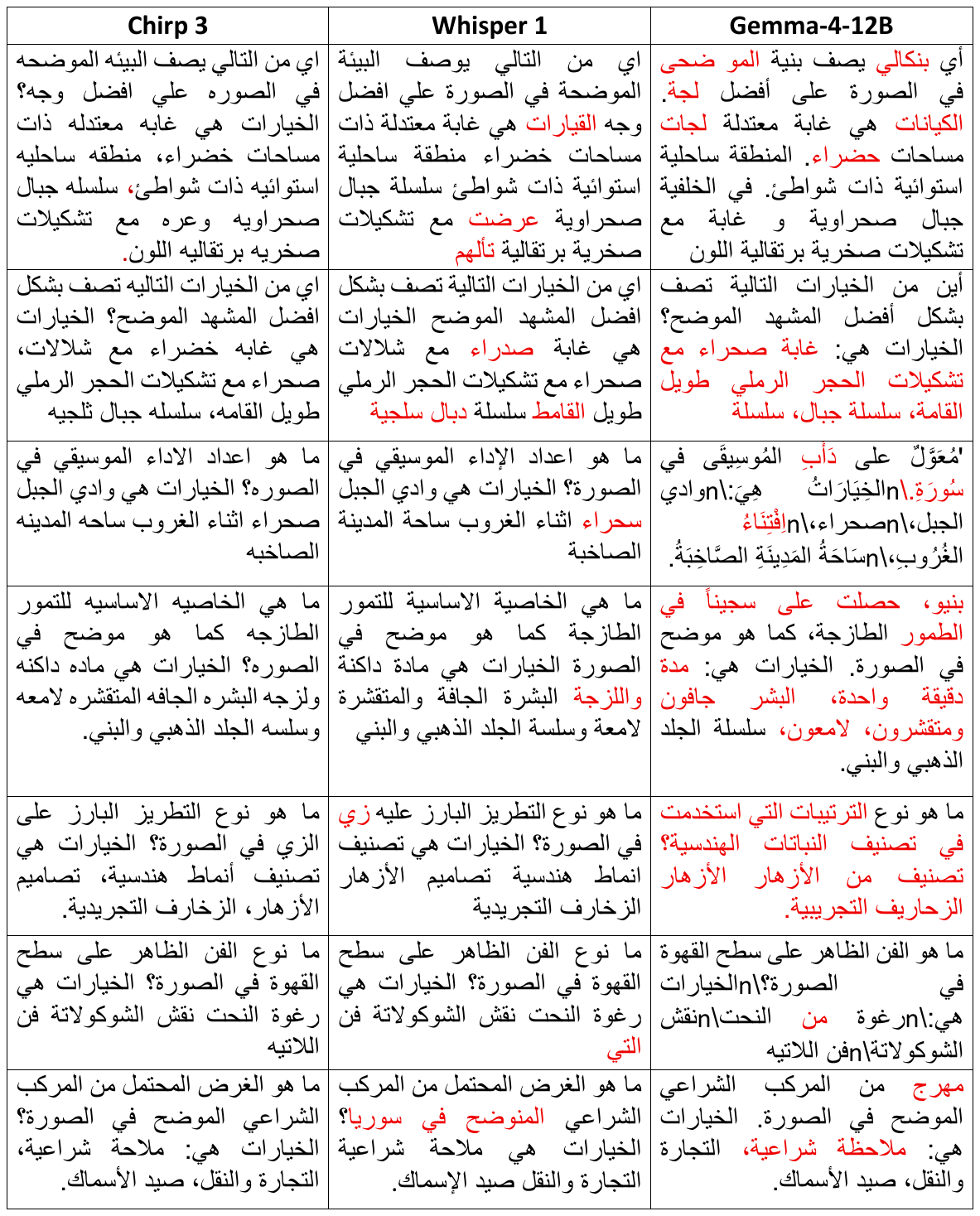}
    \caption{Examples comparing Chirp-3, Whisper-1, and Gemma-4-12B. Words in red indicate transcription errors.}
    \label{fig:compare}
\end{figure}

Figure~\ref{fig:compare} presents a qualitative comparison of MSA ASR outputs across three models, where red text marks transcription errors. A visual analysis of the results shows that Chirp 3 demonstrates the cleanest transcriptions with the fewest highlighted errors, Whisper 1 performs moderately well with occasional word substitutions and phonetic mix-ups, and Gemma-4-12B exhibits the highest error rate, frequently generating incorrect vocabulary and hallucinated phrases.

\subsubsection*{Hallucination Detections}
This section aims to answer \textbf{RQ3} and \textbf{RQ4} Table~\ref{tab:task3} and Table~\ref{tab:task4} present a comparative analysis of hallucination detection performance for multimodal tasks involving English and MSA text paired with images, demonstrating that Gemini 3.5 Flash achieves higher overall accuracy (97.2\% and 96.4\%, respectively) and completely eliminates counterfactual hallucinations, maintaining a counterfactual hallucination rate (CFHR) of zero and highlighting its stronger stability and robustness in detecting multimodal hallucinations across both languages, answering \textbf{RQ3}.

When evaluating Gemma 4-12B, its hallucination detection performance degraded substantially in MSA compared to English. In MSA, overall accuracy drops by over 10 points (from 92.0\% to 81.6\%), while its contextual instability more than doubles (jumping from 0.078 to 0.184). Additionally, its  CFHR more than triples in MSA (rising from 0.017 to 0.056), highlighting a clear gap in the model's cross-lingual robustness when handling MSA multimodal inputs, answering \textbf{RQ4}.

\vspace{-5pt}

\begin{table}[h]
\centering
\footnotesize
\begin{tabularx}{\columnwidth}{m{0.9cm}XXXXX}
\hline
\textbf{Solution} & \textbf{Cont Instab} & \textbf{Acc} & \textbf{CFHR} & \textbf{Q+Acc} & \textbf{Q-Acc} \\ \hline

\textbf{Gemma4-12B} & 0.078 & 0.920 & 0.017 & 0.936 & 0.971 \\ 

\textbf{Gemini 3.5 Flash} & \textbf{0.028} & \textbf{0.972} & \textbf{0} & \textbf{0.972} & \textbf{0.986} \\ \hline

\end{tabularx}
\caption{\normalsize Performance metrics for hallucination detection in English statements and image}
\label{tab:task3}
\end{table}

\begin{table}[h]
\centering
\footnotesize
\begin{tabularx}{\columnwidth}{m{0.9cm}XXXXX}
\hline
\textbf{Solution} & \textbf{Cont Instab} & \textbf{Acc} & \textbf{CFHR} & \textbf{Q+Acc} & \textbf{Q-Acc} \\ \hline

\textbf{Gemma4-12B} & 0.184 & 0.816 & 0.056 & 0.864 & 0.925 \\ 

\textbf{Gemini 3.5 Flash} & \textbf{0.036} & \textbf{0.964} & \textbf{0} & \textbf{0.964} & \textbf{0.982} \\ \hline

\end{tabularx}
\caption{\normalsize Performance metrics for hallucination detection in MSA statements and image}
\label{tab:task4}
\end{table}

\subsection{Final Test Phase}

\subsubsection*{Audio-Visual Question Answering }

For the final test phase using Gemini 2.5 Flash with English spoken questions, the accuracy and macro F1 score were 0.961 and  0.872, respectively. However, for MSA spoken questions using Chirp3 + Gemini 3.5 Flash, the accuracy  and macro F1 score were 0.778 and 0.632. This degradation in accuracy in MSA stems from speaker and recording conditions in the audio test set, which is human-recorded compared to synthetic (voice cloning) in the train and development sets. This issue affects MSA more than English. To improve the performance of MSA spoken visual question answering, we utilized a cascade pipeline of the Cohere-transcribe\footnote{\url{https://github.com/AliOsm/cohere-transcribe}} model as ASR + Gemini 3.5 Flash, and as a result, there was 10\%  improvement as shown in  accuracy (0.875) and macro F1 score (0.726528).

\subsubsection*{Hallucination Detections}

Using Gemini 3.5 Flash for both English and MSA hallucination detection in the final test phase, the contrastive instability, combined accuracy, CFHR, Q + Acc, and Q - Acc scores for English were 0.0410, 0.9590, 0, 0.9590, and 0.9795, respectively. For MSA, the corresponding scores were 0.0470, 0.9530, 0, 0.9530, and 0.9765, respectively. This confirms the capability of VLMs such as Gemini 3.5 Flash in detecting hallucination in VQA task.

\section*{Limitations}

In this work, we utilized pre-trained MLLMs for AVQA tasks. Native multimodal models, such as Gemini 2.5 Flash and Gemma-4-12B, exhibit degraded performance when processing spoken questions in MSA compared to English. This degradation stems from MSA's complex morphology and unique phonetic structures, such as emphatic consonants and gemination. 
Although utilizing the Gemma-4-12B model for ASR to generate text from MSA spoken questions improved accuracy by 18\%, overall accuracy remains low.
Additionally, Gemma-4 was evaluated on detecting hallucinations in MSA image-text alignments; the accuracy dropped by 10\% compared to English data.

All previous limitations suggest end-to-end fine-tuning across all input modalities, audio, vision, and language, is necessary to resolve these cross-modal alignment failures and bridge the performance gap.

\bibliography{anthology}
\bibliographystyle{acl_natbib}

\appendix

\section{Appendix}
\label{sec:appendix}

\subsection{Prompts used}

The following prompts were used in this work.

\begin{tcolorbox}[colback=green!5!white, colframe=green!75!black, title={Prompt to extract answer from pairs of transcripts and images}, rounded corners, boxrule=1pt, boxsep=1pt]
You are given:\\

* A transcript containing a multiple-choice question followed by its answer choices. The answer choices appear in order but are
**not numbered**.\\
* An image containing **one answer**.\\

Your task is to determine which answer choice in the transcript matches the answer shown in the image.\\

Return:\\

* `0` if the image matches the first answer choice.\\
* `1` if the image matches the second answer choice.\\
* `2` if the image matches the third answer choice.\\

Rules:\\

* Match based on meaning, not just exact wording. Minor differences in wording, punctuation, capitalization,
or formatting are acceptable.\\
* Do not explain your reasoning.\\
* Do not repeat the answer text.\\
* Output **only** one word: 0, 1, 2. Stop adding details.\\

Transcript:\\
{{TRANSCRIPT}}\\

Image:\\
{{IMAGE}}\\

\end{tcolorbox}

\begin{tcolorbox}[colback=purple!5!white,colframe=purple!75!black, title={Prompt to extract answers directly from pairs of spoken questions and images}, rounded corners, boxrule=1pt, boxsep=1pt]
Based on the provided image and the spoken question in the audio, please provide the correct answer in English.\\
The answer should be strictly a single number, 0, 1, or 2, based on option order.
\end{tcolorbox}

\begin{tcolorbox}[colback=blue!5!white, colframe=blue!75!black, title={Prompt to transcribe spoken questions}, rounded corners, boxrule=1pt, boxsep=1pt]
Transcribe the following speech segment in its original language.\\
Follow these specific instructions for formatting the answer:\texttt{\textbackslash n*} Only output the transcription, with no newlines.\texttt{\textbackslash n*} When
transcribing numbers, write the digits, i.e. write 1.7 and not one point seven, and write 3 instead of three.
\end{tcolorbox}

\begin{tcolorbox}[colback=blue!5!white, colframe=cyan!75!black, title={Prompt to detect hallucination in three statements}, rounded corners, boxrule=1pt, boxsep=1pt]
Based on the provided image and list of statements, please provide the correct answer. The answer should be strictly [FALSE, FALSE, TRUE] or [FALSE, TRUE, FALSE] or [TRUE, FALSE, FALSE] based on the statements' order.
\end{tcolorbox}

\subsection{Examples from Datasets}

\begin{figure}[H]
    \centering
   \includegraphics[width=1\linewidth]{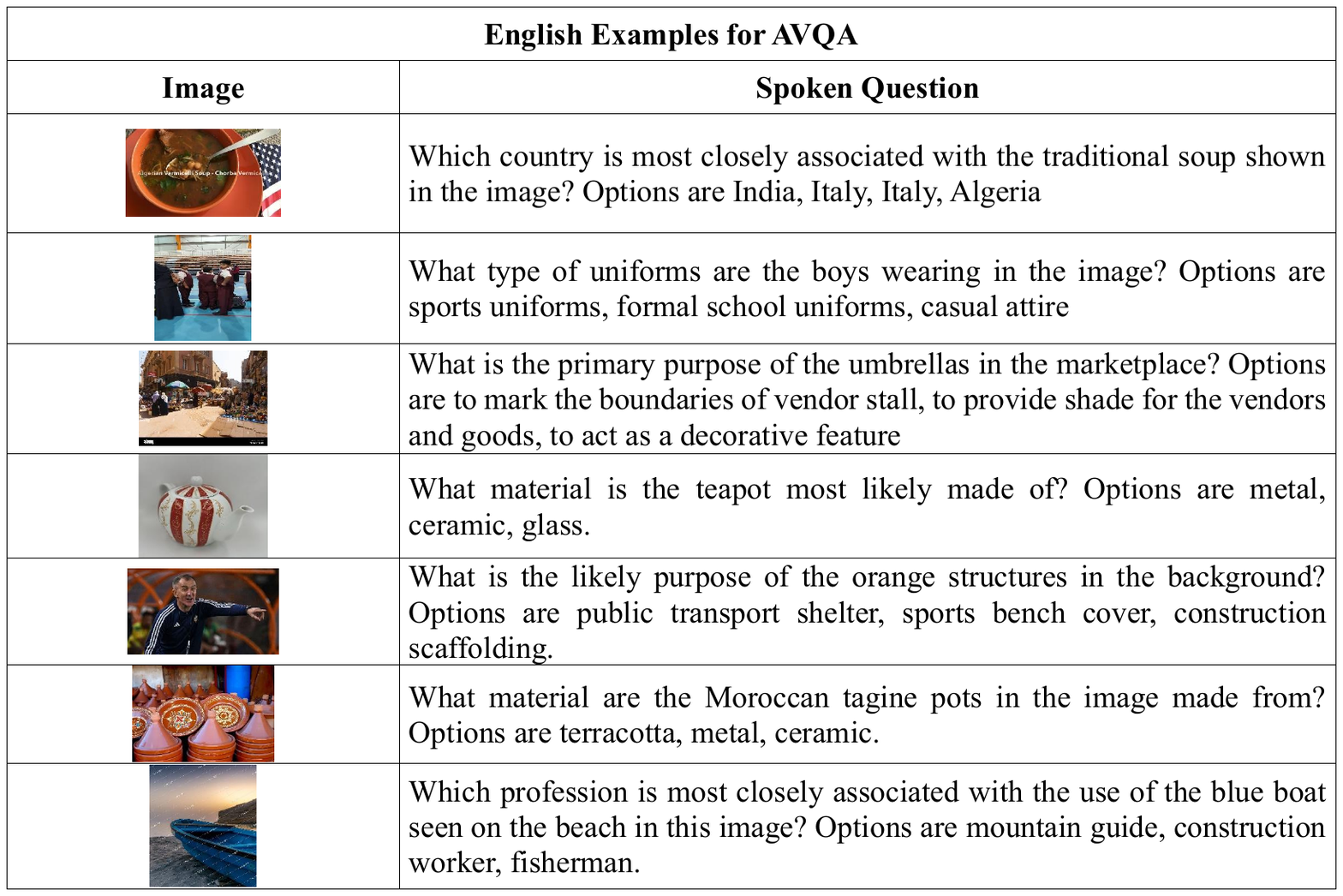}
    \caption{Examples of English spoken questions alongside images. The transcripts were generated using Chirp-3.}
    \label{fig:examples1}
\end{figure}

\begin{figure}[H]
    \centering
   \includegraphics[width=1\linewidth]{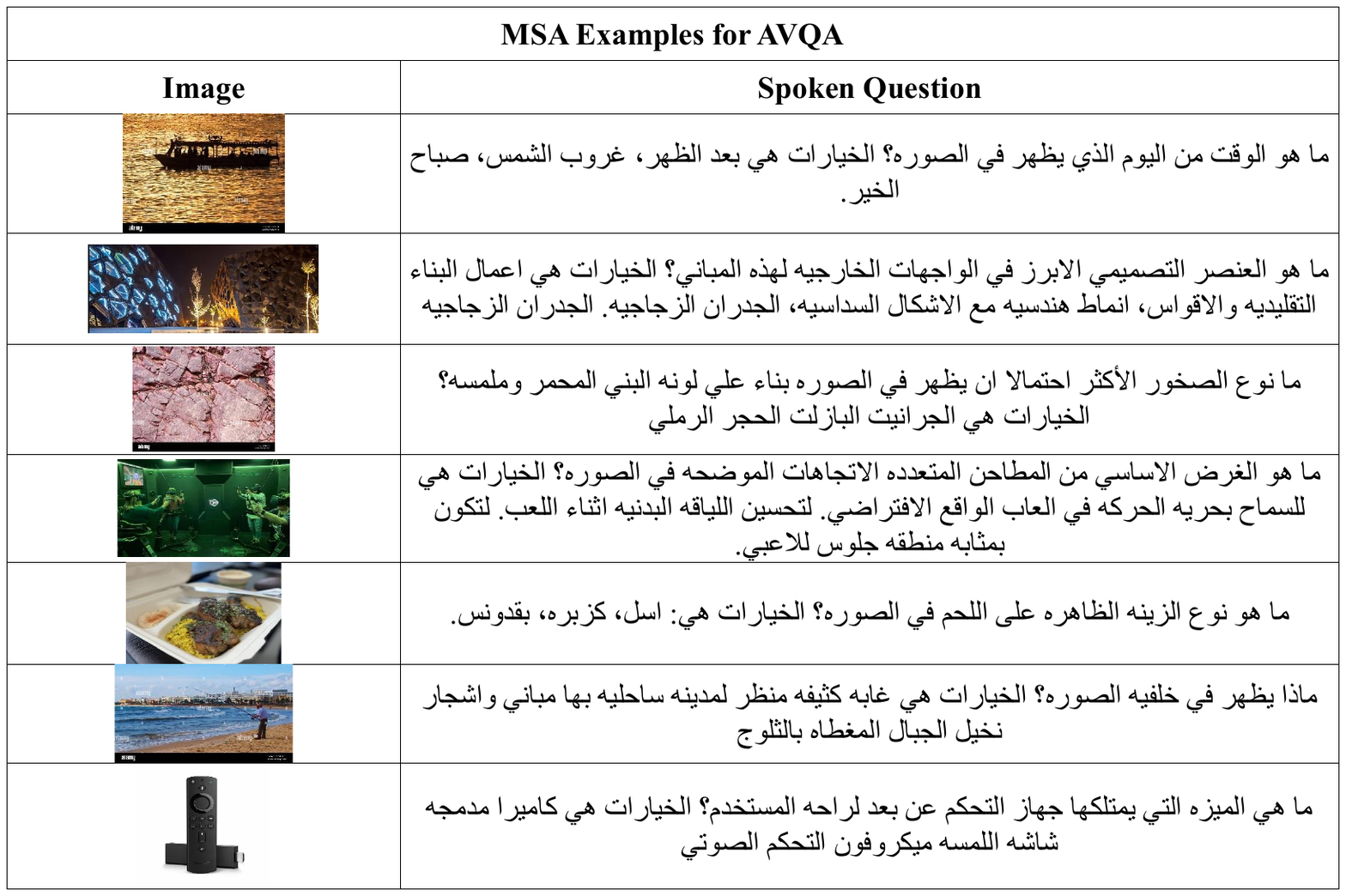}
    \caption{Examples of MSA spoken questions alongside images. The transcripts were generated using Chirp-3.}
 \label{fig:examples2}
\end{figure}

    \begin{figure}[H]
    \centering
   \includegraphics[width=1\linewidth]{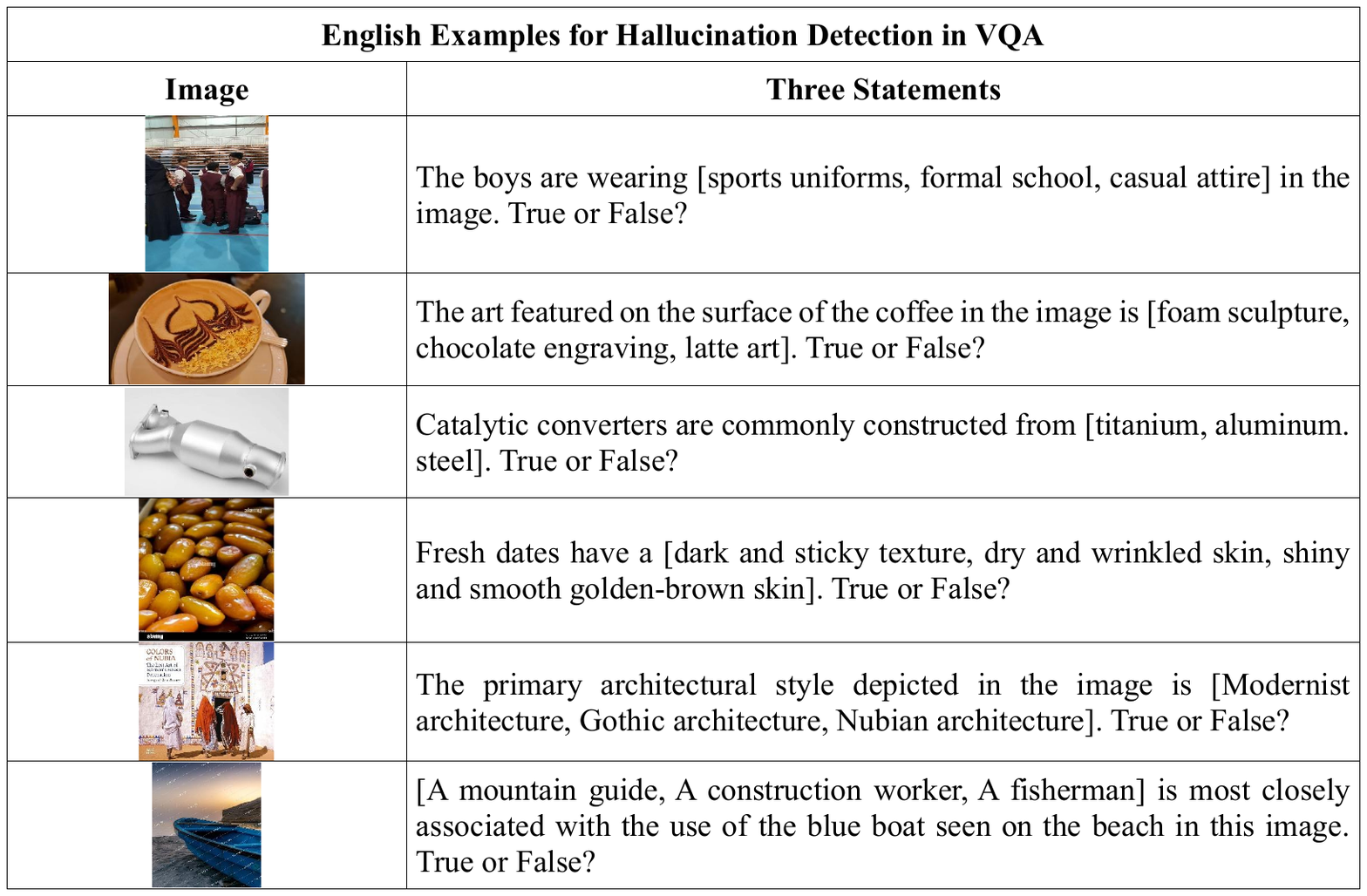}
    \caption{Examples of English statements alongside images.}
    \label{fig:examples3}
\end{figure}

\begin{figure}[H]
    \centering
   \includegraphics[width=1\linewidth]{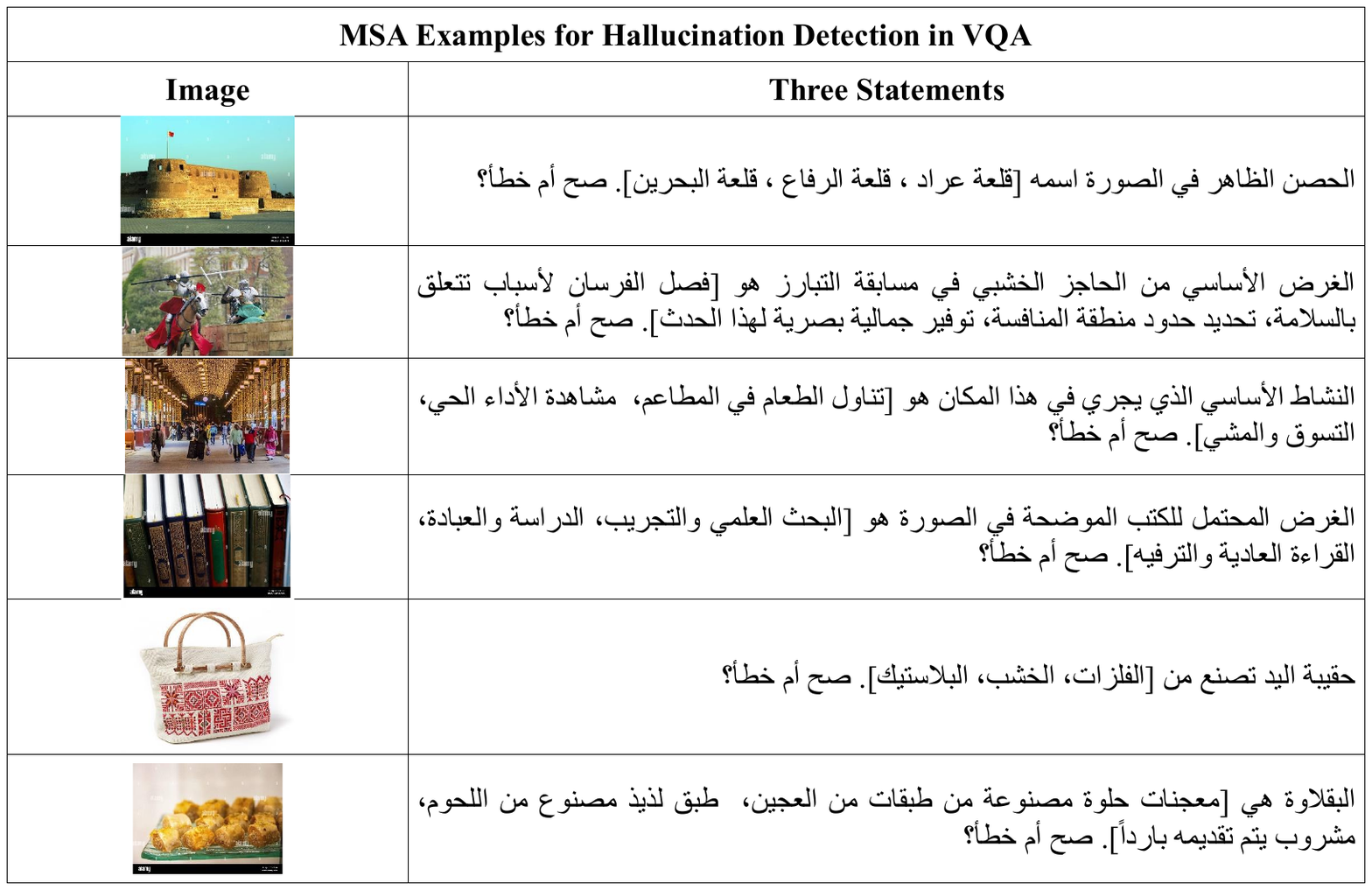}
    \caption{Examples of MSA statements alongside images.}
    \label{fig:examples4}
\end{figure}

\end{document}